\documentclass[10pt,twocolumn,letterpaper]{article}

\usepackage{iccv}
\usepackage{times}
\usepackage{epsfig}
\usepackage{graphicx}
\usepackage{amsmath}
\usepackage{amssymb}
\usepackage{multirow}
\usepackage{booktabs}
\usepackage[accsupp]{axessibility}  % Improves PDF readability for those with disabilities.
\usepackage[breaklinks=true,bookmarks=false]{hyperref}

\iccvfinalcopy % *** Uncomment this line for the final submission

\def\iccvPaperID{3709} % *** Enter the ICCV Paper ID here

\ificcvfinal\fi

\begin{document}

%%%%%%%%% TITLE
\title{Chinese Text Recognition with A Pre-Trained CLIP-Like Model Through Image-IDS Aligning}

\author{Haiyang Yu, Xiaocong Wang, Bin Li\thanks{Corresponding Author}, Xiangyang Xue\\
Shanghai Key Laboratory of Intelligent Information Processing\\ School of Computer Science, Fudan University\\
{\tt\small \{hyyu20, xcwang20, libin, xyxue\}@fudan.edu.cn}}
% For a paper whose authors are all at the same institution,
% omit the following lines up until the closing ``}''.
% Additional authors and addresses can be added with ``\and'',
% just like the second author.

\maketitle
% Remove page # from the first page of camera-ready.
\ificcvfinal\thispagestyle{empty}\fi

%%%%%%%%% ABSTRACT
\begin{abstract}
Scene text recognition has been studied for decades due to its broad applications. However, despite Chinese characters possessing different characteristics from Latin characters, such as complex inner structures and large categories, few methods have been proposed for Chinese Text Recognition (CTR). Particularly, the characteristic of large categories poses challenges in dealing with zero-shot and few-shot Chinese characters. In this paper, inspired by the way humans recognize Chinese texts, we propose a two-stage framework for CTR. Firstly, we pre-train a CLIP-like model through aligning printed character images and Ideographic Description Sequences (IDS). This pre-training stage simulates humans recognizing Chinese characters and obtains the canonical representation of each character. Subsequently, the learned representations are employed to supervise the CTR model, such that traditional single-character recognition can be improved to text-line recognition through image-IDS matching. To evaluate the effectiveness of the proposed method, we conduct extensive experiments on both Chinese character recognition (CCR) and CTR. The experimental results demonstrate that the proposed method performs best in CCR and outperforms previous methods in most scenarios of the CTR benchmark. It is worth noting that the proposed method can recognize zero-shot Chinese characters in text images without fine-tuning, whereas previous methods require fine-tuning when new classes appear. The code is available at \href{https://github.com/FudanVI/FudanOCR/tree/main/image-ids-CTR}{\textcolor{blue}{https://github.com/FudanVI/FudanOCR/tree/main/image-ids-CTR}}.
\end{abstract}

%%%%%%%%% BODY TEXT

\begin{figure}[t]
    \centering
    \includegraphics[width=0.44\textwidth]{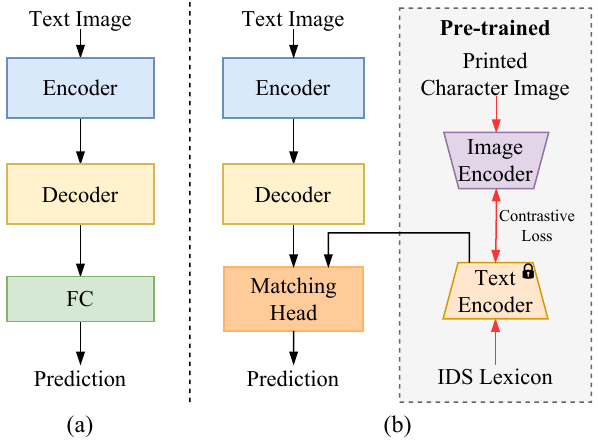}
    \caption{Comparison between the framework of previous methods (a) and that of the proposed method (b). The data flow of the pre-training stage is in red.}
    \label{fig:intro}
\end{figure}

\section{Introduction}
In recent decades, most researchers have focused on exploring Chinese character recognition (CCR)~\cite{jin2012license,ren2017novel,zhu2018scut,yu2022chinese,zu2022chinese}, few methods are dedicated to tackle Chinese Text Recognition (CTR). Unlike Latin characters, Chinese characters have a large number of categories and complex internal structures, which lead to zero-shot (\textit{i.e.}, characters in test sets are unseen in training sets) and few-shot problems in practical applications. The conventional framework for CTR should be fine-tuned with the updated alphabet when a new Chinese character appears. However, humans are able to easily match unseen character images with the corresponding characters in their stsndard (e.g. printed) forms. Thus, the question is -- \textit{Can a model recognize Chinese texts like humans}?

To tackle the zero-shot problem, existing CCR methods rely on predicting radical or stroke sequences to recognize characters. For example, some radical-based methods~\cite{wang2018denseran,wang2019radical} are proposed to decompose Chinese characters at the radical level and predict corresponding radical sequences to determine final predicted characters. Recently, a stroke-based method~\cite{chen2021zero} has been proposed to decompose Chinese characters into stroke sequences, offering a fundamental solution to the zero-shot problem in CCR. These methods are based on relatively complex networks so that they are not suitable for adoption in CTR models to solve zero-shot and few-shot problems. In addition, most scene text recognition models~\cite{li2019show,luo2019moran} adopt an encoder-decoder framework, which utilizes a fully connected layer to classify characters (as shown in Figure~\ref{fig:intro}(a)). However, these methods require to be fine-tuned when a new character appears, which is inconvenient in practical applications. Furthermore, these methods fail to account for the aforementioned unique characteristic of Chinese characters.

For native Chinese speakers, their initial learning is to recognize individual Chinese characters. In this stage, they also learn how to decompose each Chinese character into the corresponding radical sequence. When reading a text line, they first locate the position of each character and then compare it with the standard characters they have learned to determine its category. For unseen characters, people can use their knowledge of radicals and structures to deduce their categories.

Inspired by the way humans recognize Chinese texts, we propose a two-stage framework (as shown in Figure~\ref{fig:intro}(b)) to address the challenge of CTR. The proposed framework consists of a CCR-CLIP pre-training stage and a CTR stage. In the first stage, we introduce a CLIP-like model, named CCR-CLIP, to learn the canonical representations of Chinese characters through aligning printed character images and their corresponding Ideographic Description Sequences (\textit{i.e.}, radical sequences) in an embedding space. Similar to CLIP~\cite{radford2021learning}, the CCR-CLIP model comprises an image encoder and a text encoder, and is trained with a contrastive loss between embeddings of character images and embeddings of radical sequences. To ensure that the image encoder extracts features that are independent of font styles, we also introduce a contrastive loss between input images having the same label in a training batch. After pre-training, the text encoder can output the canonical representations of given radical sequences. In the CTR stage, the learned canonical representations are employed to supervise the CTR model, which is a conventional encoder-decoder framework without a fully connected layer after the decoder. During inference, the model predicts each character in a text image by calculating the similarity between the learned canonical representations and the extracted character embedding. Thus, it is able to recognize zero-shot Chinese characters without fine-tuning. We conduct extensive experiments to validate the effectiveness of the proposed method. Specifically, we train the CCR-CLIP model on several Chinese character recognition benchmarks to evaluate its performance on CCR. The experimental results show that the CCR-CLIP model can robustly recognize Chinese characters in zero-shot settings. Furthermore, our experiments on a CTR benchmark demonstrate that the proposed method outperforms previous methods in most cases.

In summary, our contributions are as follows:
\begin{itemize}
    \item Drawing inspiration from how humans recognize Chinese texts, we propose a two-stage framework for CTR, which comprises a CCR-CLIP pre-training stage and a CTR stage.
    
    \item We adopt the CLIP architecture to establish a CCR-CLIP pre-trained model to learn the canonical representations of Chinese characters.

    \item Benefiting from the learned canonical representations, the proposed method can recognize zero-shot characters in Chinese text images without fine-tuning. 
    
    \item Extensive experiments validate that the CCR-CLIP model outperforms previous CCR methods by a clear margin. Furthermore, the proposed two-stage framework for CTR achieves better performance than previous methods, particularly when training data is scarce. 
    
    % Extensive experiments show that our method surpasses state-of-the-art methods in zero-shot settings, and achieves comparable performance in non-zero-shot settings with significantly reduced inference time.
\end{itemize}

\section{Preliminaries}
% In this section, we first introduce the preliminary knowledge about Chinese characters. Then we brief introduce three categories of CCR methods, \textit{i.e.}, character-based, radical-based, and stroke-based methods.

\begin{figure}[t]
    \centering
    \includegraphics[width=0.47\textwidth]{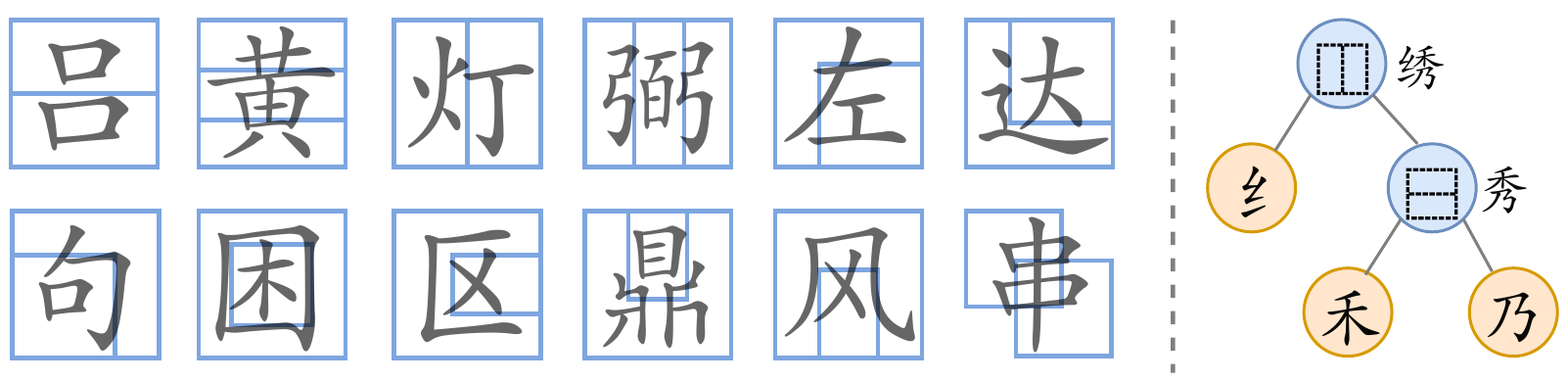}
    \caption{Twelve basic structures represented in blue lines (left) and an example of decomposition at the radical level (right).}
    \label{fig:radical}
\end{figure}
\begin{figure}[t]
    \centering
    \includegraphics[width=0.47\textwidth]{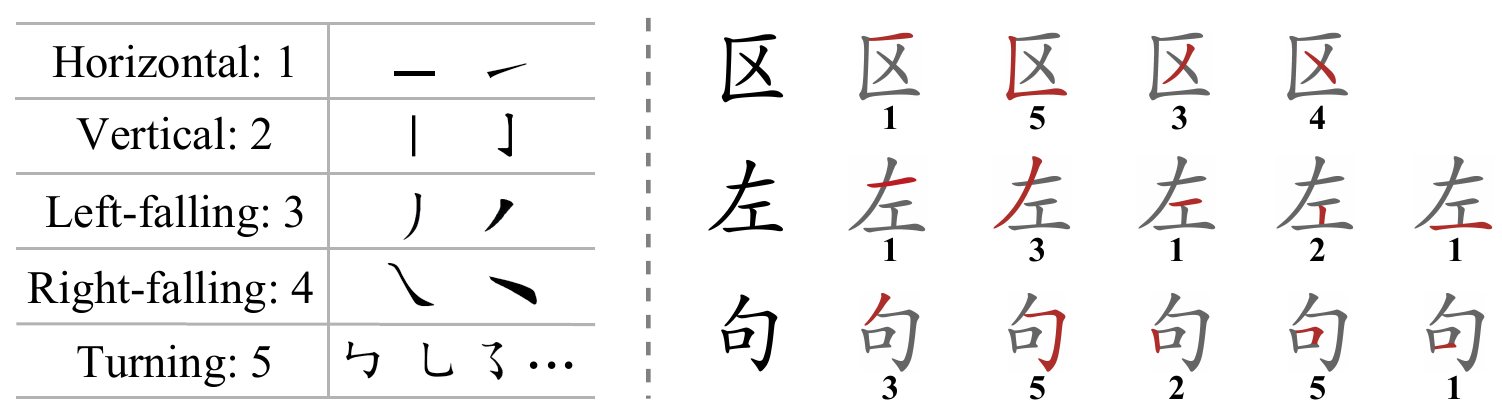}
    \caption{Five categories of strokes for Chinese characters (left) and some examples of decomposition at the stroke level (right).}
    \label{fig:stroke}
\end{figure} 

% Different from Latin characters, Chinese characters have many categories and complex inner structures. 

\subsection{Background Knowledge of Chinese Characters}
\label{decomposition}
According to Chinese national standard GB18030-2005\footnote{https://zh.wikipedia.org/wiki/GB\_18030}, there are 70,244 classes of Chinese characters, 3,755 of which are commonly-used Level-1 characters. Although Chinese characters have complex inner structures, each Chinese character can be decomposed into the corresponding radical or stroke sequence in a specific order. 
% The details of decomposition are introduced as follows.

\textbf{Radicals.} As shown in Figure~\ref{fig:radical}(right), each Chinese character can be represented as a radical tree, which can be transformed into the corresponding Ideographic Description Sequence (IDS). IDS is defined by Unicode and is composed of radicals and basic structures. Specifically, there are 514 radicals and twelve basic structures (see Figure~\ref{fig:radical}(left)) for 3,755 commonly-used Level-1 characters.

\textbf{Strokes.} There are five categories of strokes for Chinese characters according to Chinese national standard GB18030-2005. As shown in Figure~\ref{fig:stroke}(left), each category of stroke may contain several instances. Some examples of decomposing Chinese characters at the stroke level are shown in Figure~\ref{fig:stroke}(right).

\subsection{Related Work}
\textbf{Chinese Character Recognition (CCR).}
Early Chinese character recognition (CCR) methods typically rely on hand-crafted features~\cite{jin2001study,su2003novel,chang2006techniques}. With the development of deep learning, CNN-based methods like MCDNN~\cite{cirecsan2015multi} have achieved remarkable success in extracting robust features of Chinese characters, approaching human-level performance on handwritten CCR tasks in the ICDAR 2013 competition~\cite{yin2013icdar}. To address the zero-shot problem in CCR, some methods~\cite{luo2023self,li2020deep,ao2022cross,liu2022open} have been proposed to predict the radical sequences of input character images. For instance, Wang \textit{et al.}~\cite{wang2018denseran} used a DenseNet-based encoder~\cite{huang2017densely} to extract character features and an attention-based decoder to predict the corresponding radical sequence. Although such radical-based methods can partially alleviate the zero-shot problem, predicting radical sequences is more time-consuming than character-based methods. Recently, some methods attempt to decompose Chinese characters into stroke sequences to address the zero-shot problem. For example, SD~\cite{chen2021zero} decomposes each Chinese character into a sequence of strokes and employs a feature-matching strategy to address the one-to-many problem between a Chinese character and multiple stroke sequences. Although these CCR methods achieve satisfying performance on various CCR datasets, their complex structures make them unsuitable for the CTR task.

\begin{figure*}[t]
    \centering
    \includegraphics[width=1.0\textwidth]{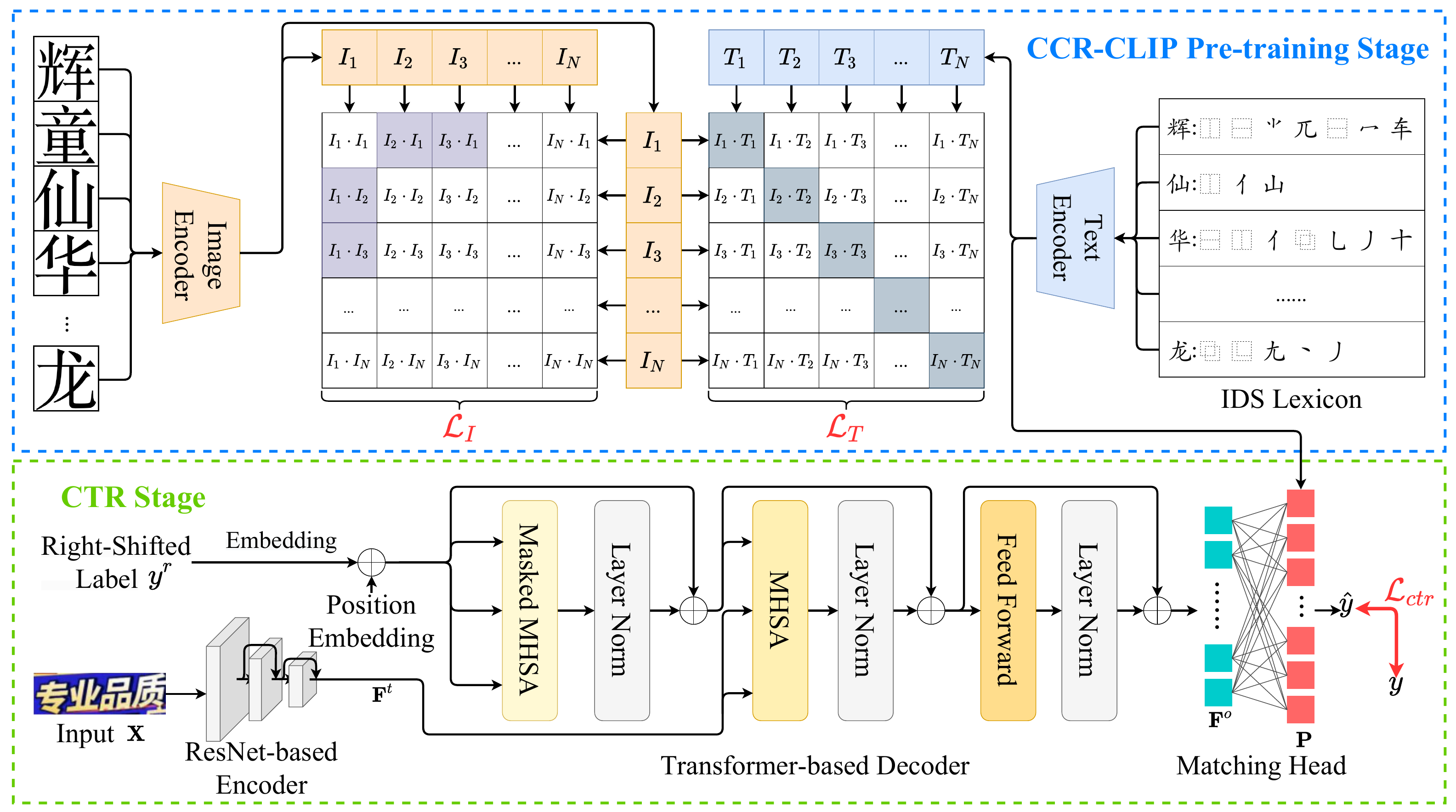}
    \caption{Overall architecture of the proposed method, consisting of a CCR-CLIP pre-training stage and a CTR stage. After being pre-trained at the pre-training stage, the CCR-CLIP model produces canonical representations of Chinese characters for the CTR model. `MHSA' represents the multi-head self-attention mechanism.}
    \label{fig:architecture}
\end{figure*}

\textbf{Chinese Text Recognition (CTR).} Scene text recognition has made significant strides in recent years. Early CTC-based text recognition methods~\cite{wan20192d,shi2016end,gao2021regularizing} tend to combine CNN and RNN to extract image features and be optimized through the CTC loss~\cite{graves2006connectionist}. To address the issue of curved texts, some methods such as ASTER~\cite{shi2018aster} and MORAN~\cite{luo2019moran} have been proposed to transform curved text images into horizontal ones. These methods have achieved promising results in curved text recognition. To incorporate semantic information into recognition models, some methods such as SEED~\cite{qiao2020seed} and ABINet~\cite{fang2021read} introduce an additional language module. Despite the impressive performance of existing methods on Latin text recognition benchmarks, CTR remains a challenging task~\cite{chen2021benchmarking}. To address this problem, a recent work~\cite{chen2021benchmarking} focuses on developing a CTR benchmark and evaluating the performance of mainstream text recognition methods. In addition, the authors proposed to introduce the radical-level supervision to improve the performance of baseline models on the CTR benchmark. However, there are still two unsolved problems: 1) These methods struggle with zero-shot and few-shot problems, which are inevitable in practical applications. 2) When a new character is supplemented in the alphabet, these models should be fine-tuned with the updated alphabet.

\section{Methodology}
In this paper, we present a novel two-stage framework for Chinese text recognition. The proposed method consists of two stages: the CCR-CLIP pre-training stage and the CTR stage. In the pre-training stage, we develop a CLIP-like model, called CCR-CLIP, which is adopted to learn canonical representations of Chinese characters. The learned representations serve as a guidance for the following CTR model. The architecture of the proposed method is depicted in Figure~\ref{fig:architecture}. Next, we provide a detailed introduction to each stage in the proposed method.

\subsection{CCR-CLIP Pre-training Stage}
Similar to CLIP~\cite{radford2021learning}, the proposed CCR-CLIP model consists of an image encoder and a text encoder. The image encoder is responsible for extracting the visual features of the input character image, while the text encoder extracts the features of the corresponding radical sequence. Finally, two contrastive losses are utilized to supervise this model. We train the CCR-CLIP model using printed Chinese character images, and the pre-trained text encoder is used to generate canonical representations for all candidate Chinese characters.

\textbf{Image Encoder.} ResNet~\cite{he2016deep} is a widely adopted feature extractor and plays a crucial role in optical character recognition tasks~\cite{yang2017improving,wang2019radical}. In the CCR-CLIP model, we use ResNet-50 to extract the feature maps $\mathbf{F}^c\in \mathbb{R}^{\frac{H}{8} \times \frac{W}{8}\times C}$ from an input printed Chinese character image. To represent the input image with a 1-D vector, we employ the global average pooling~\cite{lin2013network} to compress the feature maps $\mathbf{F}^c$:
% With the development of deep learning, ResNet~\cite{he2016deep} has been widely adopted as the feature extractor and plays a significant role in optical character recognition tasks~\cite{yang2017improving,wang2019radical}. In the CCR-CLIP pre-trained model, we use ResNet-50 to extract the feature map $\mathbf{F}^c$ of the input printed character image. To represent the input image with a 1-D vector, a global average pooling~\cite{lin2013network} is employed to compress the feature maps $\mathbf{F}^c$:
\begin{equation}
    \mathbf{f}^c = \text{GlobalAvgPool}(\mathbf{F}^c)
\end{equation}
where $\mathbf{f}^c \in \mathbb{R}^{1 \times C}$ denotes the compressed feature vector. At last, we project $\mathbf{f}^c$ into the embedded visual-feature space:
\begin{equation}
    \mathbf{I} = \mathbf{f}^c \mathbf{W}^{c}
\end{equation}
where $\mathbf{I}$ represents the embedded visual features of the input Chinese character image, $\mathbf{W}^{c} \in \mathbb{R}^{C \times C^{\prime}}$ denotes the projection matrix, and $C^{\prime}$ is the dimensionality for alignment.

\textbf{Text Encoder.} In this paper, we regard the corresponding radical sequence $\mathbf{R} = \{r_1, r_2, ..., r_l\}$ as the caption of the input Chinese character image, where $l$ denotes the length of the radical sequence and $r_l$ is an ``END'' token. The text encoder consists of $K$ layers of Transformer encoder~\cite{vaswani2017attention} and an embedding layer. Through the Transformer encoder, $\mathbf{R}$ is encoded into $\mathbf{F}^{r} = \{\mathbf{f}^{r}_1, \mathbf{f}^{r}_2, ..., \mathbf{f}^{r}_l\}$, where $\mathbf{f}^{r}_l \in \mathbb{R}^{1 \times D}$ is regarded as the whole features of $\mathbf{R}$. Similar to the image encoder, we project $\mathbf{f}^{r}_l$ into $\mathbf{T}$:
\begin{equation}
    \mathbf{T} = \mathbf{f}^{r}_l \mathbf{W}^{r}
\end{equation}
where $\mathbf{W}^{r} \in \mathbb{R}^{D \times C^{\prime}}$ denotes the projection matrix.

\textbf{Loss Function.} We employ a contrastive loss $\mathcal{L}_{T}$ to align the extracted visual features of a Chinese character image and the features of its corresponding radical sequence. For a training batch with $N$ character samples, the loss function $\mathcal{L}_{T}$ is calculated as follows:
\begin{equation}
\begin{aligned}
    \mathcal{L}_{T} = -\sum\limits^{N}_{j=1}\log \frac{\exp(\mathbf{I}_{j} \cdot \mathbf{T}_{j})}{\sum_{n=1}^N \exp(\mathbf{I}_{j} \cdot \mathbf{T}_{n})} \\ - \sum\limits^{N}_{j=1}\log \frac{\exp(\mathbf{I}_{j} \cdot \mathbf{T}_{j})}{\sum_{n=1}^N \exp(\mathbf{I}_{n} \cdot \mathbf{T}_{j})}
\end{aligned}
\end{equation}
where $\mathbf{I}_{j}$ and $\mathbf{T}_j$ represent the embedded visual features and radical sequence features of the $j$-th sample in a data batch, respectively.

To reduce the prediction errors caused by various font styles and similar characters, we additionally introduce a contrastive loss $\mathcal{L}_{I}$ between the visual features of input images having the same label in the batch. Given a data batch $\mathcal{B} = \{(\mathbf{C}_1, \mathbf{R}_1), (\mathbf{C}_2, \mathbf{R}_2), ..., (\mathbf{C}_N, \mathbf{R}_N)\}$, $\mathbf{C}_i$ and $\mathbf{R}_i$ represent the $i$-th Chinese character image and its corresponding radical sequence, respectively. Through the image encoder, the $i$-th character image $\mathbf{C}_i$ is encoded into the corresponding visual features $\mathbf{I}_i$. Thus, the loss function $\mathcal{L}_{I}$ is computed as follows:
\begin{equation}
    \mathcal{L}_{I} = -\sum\limits^{N}_{j=1}\log \frac{\sum\limits_{\mathbf{I}^{\prime} \in \mathcal{U}_{j}} \exp(\mathbf{I}_{j} \cdot \mathbf{I}^{\prime})}{\sum_{n=1}^N \exp(\mathbf{I}_{j} \cdot \mathbf{I}_{n})}
\end{equation}
where $\mathcal{U}_j$ represents the set of visual features that have the same corresponding radical sequence $\mathbf{R}_j$. Finally, the overall loss function of the CCR-CLIP model is as follows:
\begin{equation}
    \mathcal{L}_{pre} = \mathcal{L}_{T} + \lambda \mathcal{L}_{I}
\end{equation}
where $\lambda$ is the trade-off coefficient for balancing the two loss items. The experimental results of selecting $\lambda$ are shown in the Supplementary Material.

\subsection{CTR Stage}
Taking radical sequences of all candidate characters as input, the pre-trained text encoder can produce their canonical representations $\mathbf{P}= [\mathbf{p}_1, \mathbf{p}_2\, ..., \mathbf{p}_K]$, which are utilized as the supervision at the CTR stage. $\mathbf{p}_k$ denotes the canonical representation of the $k$-th candidate character and $K$ is the number of candidate characters. For the CTR model, we adopt a conventional encoder-decoder framework that consists of a ResNet-based encoder, a Transformer-based decoder, and a matching head.

\textbf{ResNet-based Encoder.} Given the input text image $\mathbf{X}$, the ResNet-based encoder is employed to extract its visual features $\mathbf{F}^t$. We modify some layers in the original ResNet-34. First, we replace the $7\times7$ kernel of the first convolution layer with a $3\times3$ kernel since the smaller kernel can capture more details for recognizing text images. Additionally, we remove the last convolution block to reduce the number of parameters in the encoder, thereby improving the efficiency of feature extraction. Finally, we remove the max pooling layer of the third convolution block in ResNet-34 to reserve more visual features for the subsequent decoder.

\textbf{Transformer-based Decoder.} As shown in Figure~\ref{fig:architecture}, the Transformer-based decoder consists of three modules: the masked multi-head self-attention (MHSA) module, the MHSA module, and the feed-forward module. The masked MHSA module takes the right-shifted ground truth $y^r$ as input and captures the semantic dependence between characters. The MHSA module calculates the attention weights between the extracted visual features $\mathbf{F}^t$ and $y^r$. Finally, the weighted features are fed into the feed-forward module to extract deeper features $\mathbf{F}^o \in \mathbb{R}^{T \times C}$, where $T$ indicates the length of the text, and $\mathbf{F}^o_i \in \mathbb{R}^{1 \times C}$ represents the feature of the $i$-th character in the input text image.

\textbf{Matching Head.} The previous methods~\cite{shi2018aster,lu2021master} simply utilize a prediction head, \textit{i.e.}, a fully connected layer, to generate the final prediction $\hat{y} = \text{Softmax}(\mathbf{W}^t \mathbf{F}^o + \mathbf{b})$, where $\mathbf{W}^t$ and $\mathbf{b}$ represent the linear transformation and the bias of the prediction head, respectively. Different from these methods, we use the canonical representations of candidate characters $\mathbf{P}$ to match the features of input text image $\mathbf{F}^o$. Thus, the final prediction are generated by:
\begin{equation}
    \hat{y} = \text{Softmax}(\mathbf{P} \mathbf{F}^o)
\end{equation}

\textbf{Loss Function.} The learning objective supervising the CTR model contains two terms:
\begin{equation}
    \mathcal{L}_{ctr} = \sum_{\mathbf{f} \in \mathbf{F}^o} (-\log p(y|\mathbf{f}) + \beta R(\mathbf{p}_y, \mathbf{f}))
\end{equation}
where $-\log p(y|\mathbf{f})$ is the cross-entropy loss, $R(\mathbf{p}_y, \mathbf{f})$ represents the regularization term, and $\beta$ is a hyperparameter to balance these two terms. The experiment for choosing $\beta$ is shown in the Supplementary Material. $p(y|\mathbf{f})$ is calculated by:
\begin{equation}
    p(y|\mathbf{f}) = \frac{\exp(\mathbf{p}_y \cdot \mathbf{f})}{\sum\limits_{\mathbf{p}_i \in \mathbf{P}} \exp(\mathbf{p}_{i} \cdot \mathbf{f})}
\end{equation}
As shown in~\cite{yang2018robust}, a regularization term is introduced to avoid overfitting on seen classes, which is defined as:
\begin{equation}
    R(\mathbf{p}_y, \mathbf{f}) = ||\mathbf{p}_y - \mathbf{f}||^2_2
\end{equation}

\begin{figure}[t]
    \centering
    \includegraphics[width=0.47\textwidth]{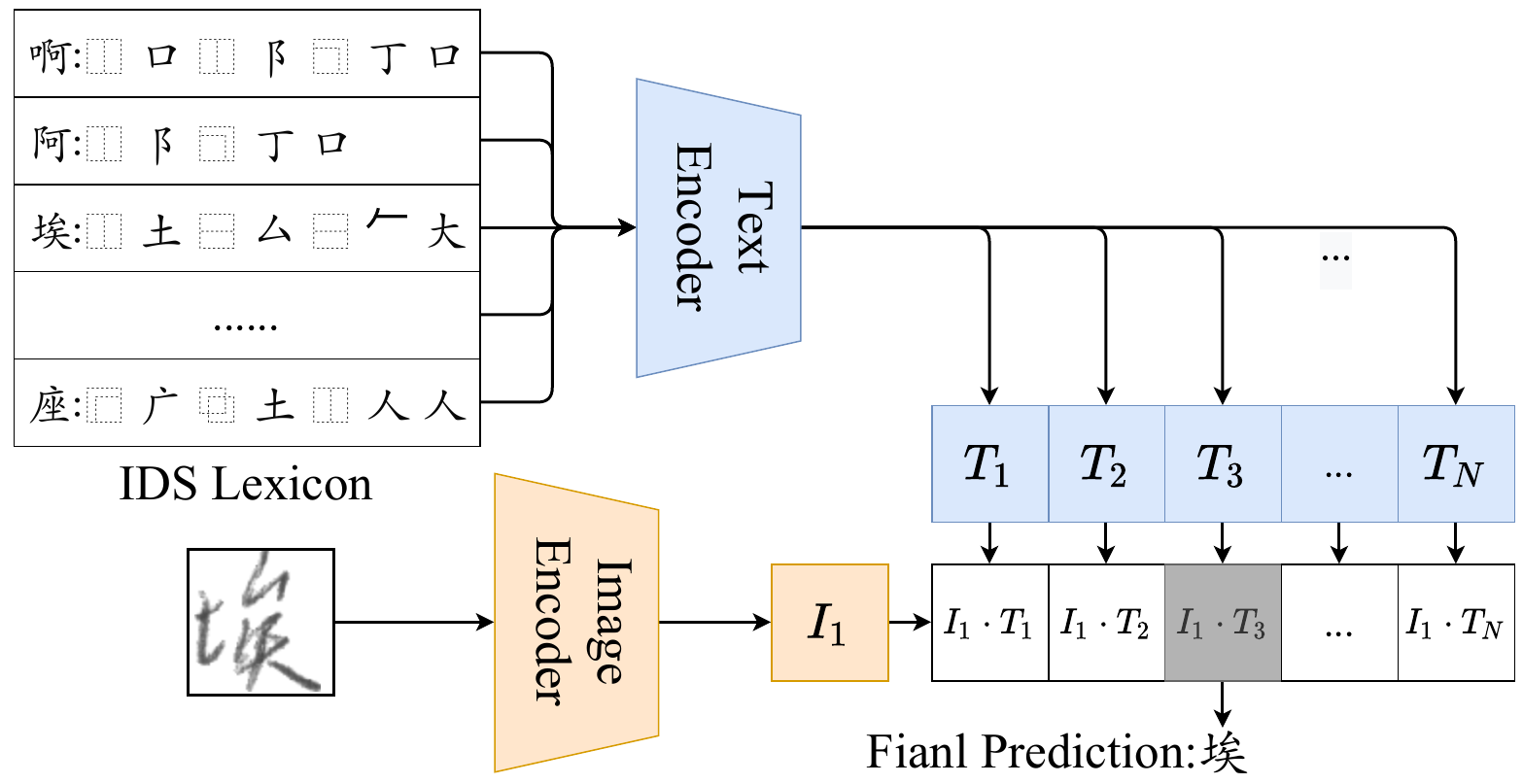}
    \caption{The test process of the CCR-CLIP model for Chinese character recognition.}
    \label{fig:test}
\end{figure}

% \begin{figure}[t]
%     \centering
%     \includegraphics[width=0.45\textwidth]{dataset.pdf}
%     \caption{Examples of the adopted datasets.}
%     %  The most samples in the handwritten dataset HWDB1.0-1.1 and ICDAR2013 have join-up strokes. The character images in CTW tend to involve the occlusions and blurring while the printed artistic character images have various font styles.
%     \label{fig:datasets}
% \end{figure}

\section{Experiments}
% In this section, we first introduce the adopted character datasets, the evaluation metric, and implementation details. Then we show the results of conducted experiments. Finally, we discuss the choices of hyperparameters.

\textbf{Datasets.} Extensive experiments on both CCR and CTR are conducted to validate the effectiveness of the proposed method. The adopted datasets are introduced in the following. Examples of each dataset are shown in the Supplementary Material.
\begin{itemize}
    \item \textbf{HWDB1.0-1.1}~\cite{liu2013online} contains 2,678,424 handwritten Chinese character images with 3,881 classes. This dataset is collected from 720 writers and covers 3,755 commonly-used Level-1 Chinese characters.
    
    \item \textbf{ICDAR2013}~\cite{yin2013icdar} contains 224,419 handwritten Chinese character images with 3,755 classes, which are collected from 60 writers.
    
    % \item \textbf{Printed artistic characters}~\cite{chen2021zero} are generated in 105 font files and contains 394,275 samples for 3,755 Level-1 Chinese characters. 
    
    \item \textbf{CTW}~\cite{yuan2019large} is collected from street views, containing 812,872 Chinese character images with 3,650 classes, where 760,107 character images are used for training and 52,765 images are used for testing. This dataset is more challenging due to its complex backgrounds and various fonts.

    \item \textbf{CTR Benchmark}~\cite{chen2021benchmarking} collects four types of Chinese text recognition datasets including scene, web, document, and handwriting. Training, validation and test datasets are divided for each type. In this paper, to fully explore the performance on Chinese texts, we filter out those samples containing non-Chinese characters. Details about these four adopted datasets are introduced in the Supplementary Material.
\end{itemize}

\textbf{Evaluation Metrics.} Following the previous CCR works~\cite{wang2018denseran,cao2020zero,zhang2020radical,xiao2019template}, we select Character ACCuracy (CACC) as the evaluation metric for CCR. We follow~\cite{chen2021benchmarking} to adopt two mainstream metrics to evaluate our method in CTR: Line ACCuracy (LACC) and Normalized Edit Distance (NED). LACC is defined as:
\begin{equation}
    \text{LACC} = \frac{1}{S} \sum^S_{i=1}\mathbb{I}(\hat{\mathbf{y}}_i=\mathbf{y}_i)
\end{equation}
where $S$ is the number of text images; $\mathbb{I}$ denotes the indicator function; $\hat{\mathbf{y}}_i$ and $\mathbf{y}_i$ denote the prediction and the label of the $i$-th text image, respectively. NED is defined as:
\begin{equation}
    \text{NED} = 1 - \frac{1}{S} \sum^S_{i=1} \text{ED}(\hat{\mathbf{y}}_i, \mathbf{y}_i)/\text{Maxlen}(\hat{\mathbf{y}}_i, \mathbf{y}_i)
\end{equation}
where ``ED'' and ``Maxlen'' denote the edit distance and the maximum sequence length, respectively.

\textbf{Implementation Details.} Our method is implemented with PyTorch, and all experiments are conducted on an NVIDIA RTX 4090 GPU with 24GB memory. The Adam optimizer is adopted to train the model with an initial learning rate $10^{-4}$, and the momentums $\beta_1$ and $\beta_2$ are set to 0.9 and 0.98, respectively. The batch size is set to 128. For fair comparison with previous methods, the input sizes for CCR and CTR are $32 \times 32$ and $32 \times 256$, respectively. In the text encoder, the number of Transformer encoder layers is empirically set to 12.

\begin{table}[tb]
\renewcommand{\arraystretch}{1.2}
%\begin{subtable}

\centering
\scalebox{0.8}{
\begin{tabular}{l|ccccc }
\hline 
\multirow{2}*{\textbf{HWDB}} &  \multicolumn{5}{c}{$m$ for character Zero-Shot Setting}\\
\cline{2-6}
~ & 500 & 1000 & 1500 & 2000 & 2755\\ 
\hline
DenseRAN~\cite{wang2018denseran} & 1.70\% & 8.44\% & 14.71\% & 19.51\% & 30.68\%\\
HDE~\cite{cao2020zero} & 4.90\% & 12.77\% & 19.25\% & 25.13\% & 33.49\% \\
SD~\cite{chen2021zero} & 5.60\% & 13.85\% & 22.88\% & 25.73\% & 37.91\% \\
CUE~\cite{luo2023self} & 7.43\% & 15.75\% & 24.01\% & 27.04\% & 40.55\% \\
Ours & \textbf{21.79\%} & \textbf{42.99\%} & \textbf{55.86\%} & \textbf{62.99\%} & \textbf{72.98\%} \\
\hline
DMN~\cite{li2020deep} & 66.33\% & 79.09\% & 84.14\% & 86.79\% & 88.98\%\\
CMPL~\cite{ao2022cross} & 72.49\% & 80.57\% & 84.40\% & 86.47\% & 89.29\%\\
CCD~\cite{liu2022open} & 90.93\% & 94.10\% & 94.58\% & 95.55\% & - \\
Ours & \textbf{93.80\%} & \textbf{94.97\%} & \textbf{95.35\%} & \textbf{95.71\%} & \textbf{95.73\%} \\
\hline
\end{tabular}}
\caption{Results in character zero-shot settings. $m$ represents that samples of the first $m$ classes are used for training in CCR zero-shot settings. The results in the top row are only based on HWDB while the results in the bottom row are obtained with additional printed character images during training.}
\label{big-table}
\end{table}

\begin{table}[t]
\centering
\renewcommand{\arraystretch}{1.1}
%\begin{subtable}
\scalebox{0.9}{
\begin{tabular}{lccc}
\hline 
Method & ICDAR2013 & CTW & AIT (ms) \\
\hline
ResNet~\cite{he2016deep} & 96.83\% & 79.46\% & \textbf{12} \\
DenseNet~\cite{huang2017densely} & 95.90\% & 79.88\% & 89 \\
DenseRAN~\cite{wang2018denseran} & 96.66\% & 85.56\% & 1666 \\
FewshotRAN~\cite{wang2019radical} & 96.97\% & 86.78\% & 83 \\
Template+Instance\cite{xiao2019template}$^*$ & \textit{97.45}\% & - & - \\
RAN~\cite{zhang2020radical} & 93.79\% & 81.80\% & 117 \\
HDE~\cite{cao2020zero} & 97.14\% & \textbf{89.25\%} & 29 \\
SD~\cite{chen2021zero} & 96.28\% & 85.29\% & 567 \\
\hline
Ours & \textbf{97.18\%} & 85.78\% & 14 \\
\hline
\end{tabular}}
\caption{Comparison of performance and average inference time (AIT). Methods marked with `*' use additional template character images at the training stage.}
\label{ICDAR2013}
\end{table}

\subsection{Results on Chinese Character Recognition}
Although the primary objective of the CCR-CLIP model is to generate canonical representations of Chinese characters through aligning printed character images and IDS, it also has the potential to be adapted to recognize Chinese character images through image-IDS matching. The process of inference is depicted in Figure~\ref{fig:test}.

\textbf{Experiments in Zero-shot Settings.} Due to the significantly larger alphabet size of Chinese characters, the zero-shot problem is inevitable in practical applications. To address this problem, we follow the approach of~\cite{chen2021zero} and construct corresponding datasets for character zero-shot settings. Specifically, we collect samples with labels falling in the first $m$ classes to form the training set, and collect those in the last $k$ classes for testing. For the handwritten character dataset HWDB, $m$ ranges in \{500, 1000, 1500, 2000, 2755\}, and $k$ is set to 1000.
% Since the alphabet size of Chinese characters is much larger than that of Latin characters, it is inevitable to encounter the zero-shot problem in practical applications. We follow~\cite{chen2021zero} to construct the corresponding datasets for the character zero-shot settings. For the character zero-shot settings, we collect samples with labels falling in the first $m$ classes as the training set and the last $k$ classes as the test set. In the handwritten character dataset HWDB, $m$ ranges in \{500, 1000, 1500, 2000, 2755\}, and $k$ is set to 1000.

The experimental results reported in Table~\ref{big-table} are grouped according to whether printed character images are utilized during the training stage. In the setting of no printed character images used for training, the CCR-CLIP model achieves an improvement of 28.37\% in average for the character zero-shot settings, compared with CUE~\cite{luo2023self}. These results demonstrate the effectiveness of the proposed method. Furthermore, we also incorporate additional printed character images during training, following the approach of~\cite{li2020deep}. The experimental results show that the proposed CCR-CLIP model still outperforms the compared methods in all character zero-shot settings. The additional experimental results on other datasets and radical zero-shot settings are reported in the Supplementary Material.

% The experimental results are shown in Table~\ref{big-table}. Compared methods are divided into two groups according to whether utilizing printed character images at the training stage. Compared with those methods without printed character images during training, the CCR-CLIP model outperforms previous methods by a large margin (average 30.13\% improvement) in the character zero-shot settings, which validates the effectiveness of our method. Further, we use additional printed character images during training as in~\cite{li2020deep}. The experimental results demonstrate that the proposed CCR-CLIP model can still achieve the best performance in all character zero-shot settings. More experimental results on other datasets and in radical zero-shot settings are shown in the Supplementary Material.

% The above experiments are all based on the input size $32 \times 32$. Empirically, increasing the input size of character images will improve the performance and also need more computation~\cite{li2018building}. Due to the simple architecture of CCR-CLIP model, we attempt to resize input images with a larger input size to further improve the performance of our method. The experimental results demonstrate that when input images are resized to $128 \times 128$, the performance of our method can be further improved (average 3.08\% improvement in the handwritten character zero-shot settings). More results are shown in the Supplementary Material.

\begin{figure}[t]
    \centering\includegraphics[width=0.5\textwidth]{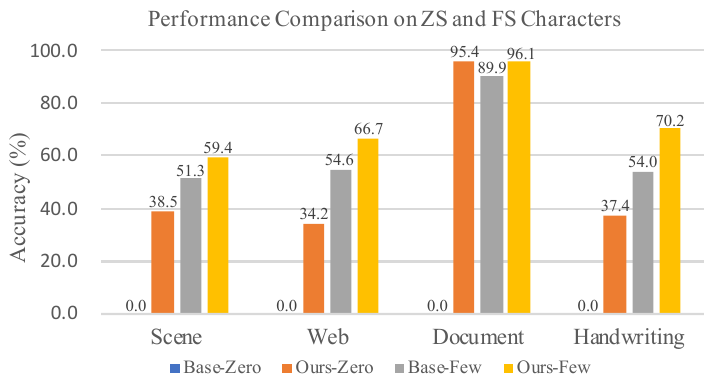}
    \caption{Performance comparison between the baseline model TransOCR and the proposed method in recognizing the zero-shot (ZS) and few-shot (FS) Chinese characters.}
    \label{fig:zs-fs}
\end{figure}

\begin{table*}[t]
\centering
\scalebox{0.85}{
\begin{tabular}{l ccccc}
\toprule
\multirow{2}*{Method} & \multicolumn{4}{c}{Dataset} & \multirow{2}*{Average} \\
\cmidrule{2-5}
~  & Scene & Web & Document & Handwriting \\
\midrule
CRNN~\cite{shi2016end} & 53.41 / 0.712 & 57.00 / 0.716 & 96.62 / 0.992 & 50.83 / 0.814 & 63.66 / 0.792 \\
ASTER~\cite{shi2018aster} & 61.34 / 0.815 & 51.67 / 0.715 & 96.19 / 0.991 & 37.00 / 0.683 & 65.69 / 0.836  \\
MORAN~\cite{luo2019moran} & 54.61 / 0.684 & 31.47 / 0.446 & 86.10 / 0.962 & 16.24 / 0.305 & 55.26 / 0.682 \\ 
SAR~\cite{li2019show} & 59.67 / 0.766 & 58.03 / 0.716 & 95.67 / 0.988 & 36.49 / 0.736 & 65.07 / 0.811\\
SEED~\cite{qiao2020seed} & 44.72 / 0.681 & 28.06 / 0.460 & 91.38 / 0.980 & 20.97 / 0.475 & 51.43 / 0.626  \\
MASTER~\cite{lu2021master} & 62.82 / 0.726 &	52.05 / 0.620 & 84.39 / 0.944 & 26.92 / 0.443 & 62.39 / 0.773 \\
ABINet~\cite{fang2021read} & 66.55 / 0.792 & 63.17 / 0.776 & 98.19 / 0.996 & 53.09 / 0.813 & 72.06 / 0.847 \\
TransOCR~\cite{chen2021scene} & 71.33 / 0.823 & 64.81 / 0.764 & 97.07 / 0.993 & 53.00 / 0.797 & 74.55 / 0.843 \\
TransOCR + PRAB~\cite{chen2021scene}  & \textbf{71.60} / \textbf{0.834} & 65.52 / 0.782 & 97.36 / 0.994 & 53.67 / 0.802 & 74.91 / 0.852 \\
\midrule
Ours & 71.31 / 0.829 & \textbf{69.21} / \textbf{0.797} & \textbf{98.29} / \textbf{0.997} & \textbf{60.30} / \textbf{0.849} & \textbf{76.13} / \textbf{0.892} \\
\bottomrule
\end{tabular}}
\caption{Comparison with previous methods on the CTR benchmark. LACC / NED follows the percentage and decimal format, respectively.
}
\label{CTR-res}
\end{table*}

\begin{figure*}[t]
    \centering\includegraphics[width=1.0\textwidth]{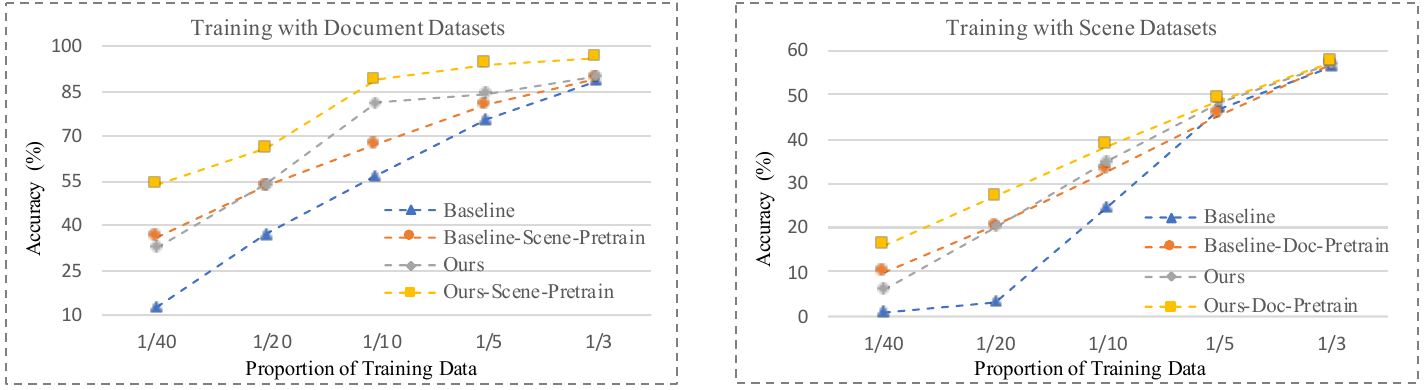}
    \caption{Performance comparison in data-scarce situations. ``Scene-Pretrain'' and ``Doc-Pretrain'' indicate that the model is pre-trained on the scene and document datasets, respectively. The proposed method performs better when the same strategy is adopted.}
    \label{fig:part-of-training}
\end{figure*}

\textbf{Experiments in Non-zero-shot Settings.} In contrast to zero-shot settings, we train the proposed CCR-CLIP model using all training samples in non-zero-shot settings, where all characters in the test dataset are covered by the training dataset. For handwritten characters, we use HWDB1.0-1.1 as the training set and ICDAR2013 as the test set. The experimental results reported in Table~\ref{ICDAR2013} show that the proposed method achieves the second-best performance, trailing only the template-instance method~\cite{xiao2019template} that benefits from additional template character images during training. Moreover, the CCR-CLIP model outperforms radical-based methods~\cite{cao2020zero,wang2018denseran,wang2019radical} with less inference time. However, the experimental results obtained on the scene character dataset CTW suggest that there is still much room for existing methods to further improve in performance, as the samples in CTW often suffer from severe occlusion and blurring problems, which indeed poses difficulties to CCR methods. We also conduct experiments to evaluate the time efficiency of the proposed method for a comprehensive comparison with existing methods. To ensure fairness, we set the batch size to 32 and calculate the average inference time for 200 batches during the test stage. As shown in Table~\ref{ICDAR2013}, the CCR-CLIP method exhibits higher time efficiency than decomposition-based CCR methods.

% Different from zero-shot settings, we use all training samples to train the proposed CCR-CLIP model in non-zero-shot settings, where all characters in the test dataset are covered by the training dataset. For handwritten characters, we use HWDB1.0-1.1 as the training set and ICDAR2013 as the test set. The experimental results (shown in Table~\ref{ICDAR2013}) demonstrate that the performance of our method is second only to that of the template-instance method~\cite{xiao2019template} that benefits from the additional template character images at the training stage. Moreover, the CCR-CLIP model outperforms the compared radical-based methods~\cite{cao2020zero,wang2018denseran,wang2019radical} with less inference time. The results of experiments conducted on the scene character dataset CTW demonstrate that the performance of our method still has room for further improvement since samples in this dataset usually have severe occlusions and blurring, which indeed brings difficulties for existing CCR methods. Moreover, we conduct experiments to evaluate the time efficiency for comprehensive comparison with previous CCR methods. For fair comparison, we set the batch size to 32. The average inference time for 200 batches in the test stage is used for comparison. As shown in Table~\ref{ICDAR2013}, the CCR-CLIP pre-trained method has higher time efficiency than other decomposition-based CCR methods.

\begin{figure*}[t]
    \centering
    \includegraphics[width=0.95\textwidth]{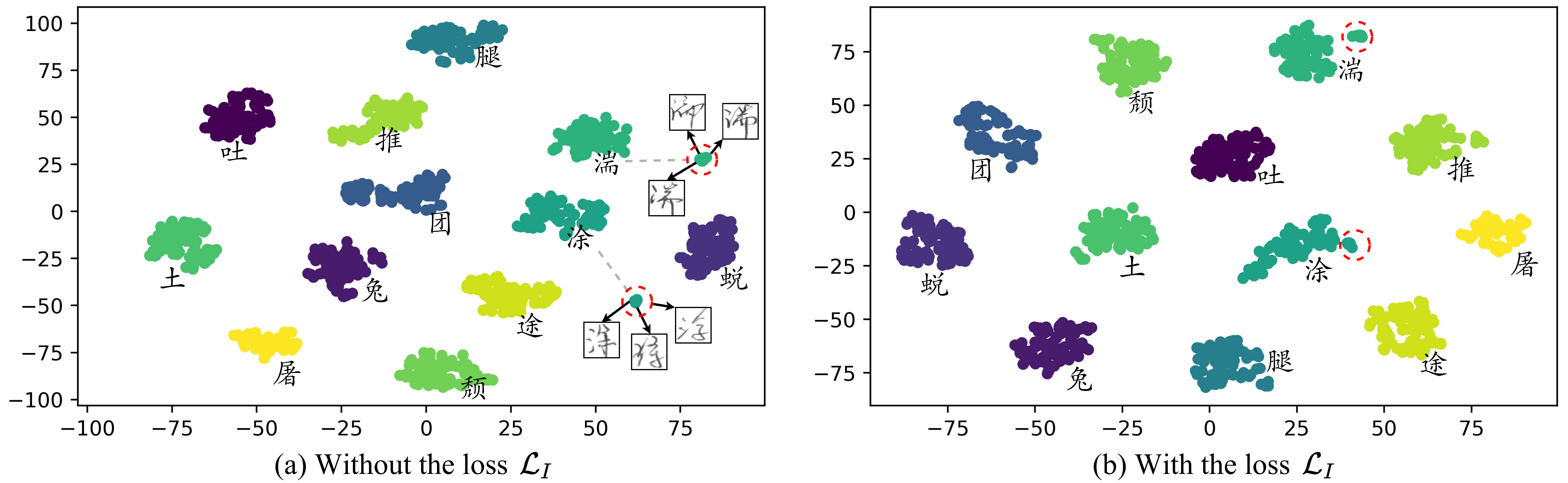}
   \caption{Character distribution visualization of whether introducing the loss $\mathcal{L}_I$ into the proposed CCR-CLIP model. The samples in red circles in (a) represent outliers with incorrect predictions, and the corresponding samples are also marked by red circles in (b). The gray lines connect the outliers and the class centers that they should correspond to.}
    \label{fig:img-loss}
\end{figure*}

\subsection{Results on Chinese Text Recognition}
We conduct experiments on a recently proposed benchmark for Chinese text recognition~\cite{chen2021benchmarking}, which contains four types of datasets: scene, web, document, and handwriting. The experimental results reported in Table~\ref{CTR-res} demonstrate that the proposed two-stage CTR method outperforms previous methods by a clear margin on the web, document, and handwriting datasets. We also evaluate the recognition accuracy of zero-shot and few-shot (1-50 shots) characters on the test sets of four types (see Figure~\ref{fig:zs-fs}). Benefiting from the design of the proposed image-IDS matching framework, our method can easily recognize zero-shot and few-shot Chinese characters. The results indicate that the proposed method performs much better than the baseline model TransOCR~\cite{chen2021scene} in both zero-shot and few-shot character settings. Specifically, our method achieves an accuracy of 51.4\% in average in the zero-shot character setting, while TransOCR cannot recognize them at all. In the few-shot character setting, the proposed method achieves an improvement of 10.6\% in average across the four types of datasets. However, we observe that the performance of our method is subpar on the scene dataset. A possible reason is that around 1/5 of the samples in the training set are vertical, which poses difficulties for our method.

% We conduct experiments on a recently proposed Chinese text recognition benchmark~\cite{chen2021benchmarking}, including four scenarios: scene, web, document, and handwriting. The experimental results shown in Table~\ref{CTR-res} demonstrate that the proposed prototype matching CTR method outperforms previous methods by a clear margin in the web, document, and handwriting datasets, resulting from that the framework design of our method makes it easy to recognize zero-shot and few-shot Chinese characters. The accuracy of zero-shot and few-shot (1-50 shots) character in the test datasets of four scenarios is shown in Figure~\ref{fig:zs-fs}. The results indicate that the proposed method performs better on both zero-shot and few-shot characters. Especially, the proposed method achieves average 51.4\% accuracy on zero-shot characters while the baseline model TransOCR~\cite{chen2021scene} cannot recognize them at all. For few-shot characters, the proposed prototype matching method gains average 10.6\% improvement on four types of datasets. However, we observe that the performance of our method is subpar in the scene datasets. A possible reason is that about 1/5 samples in the training dataset are vertical, which brings difficulties for our method.

In practical applications, collecting a large amount of annotated training data for the target domain is difficult and time-consuming. To further explore the effectiveness of our method in the case of limited training data, we randomly select subsets from the training data of the scene and document types. As shown in Figure~\ref{fig:part-of-training}, when using the same training strategy, our method outperforms the baseline model TransOCR by a clear margin on both scene and  document datasets. This validates the robustness of our method in data-scarce situations.

\begin{table}[t]
\renewcommand{\arraystretch}{1.1}

\centering
\scalebox{0.80}{
\begin{tabular}{cc|cccc}
\hline 
MH & RT & Scene & Web & Document & Handwriting \\
\hline
 & & \textbf{71.33\%} & 64.81\% & 97.07\% & 53.00\% \\
 \checkmark & & 70.17\% & 67.95\% & 97.97\% & 58.54\% \\
 \checkmark & \checkmark & 71.31\% & \textbf{69.21\%} & \textbf{98.29\%} & \textbf{60.30\%}\\
\hline
\end{tabular}}
\caption{Results of ablation study. ``MH'' and ``RT'' denote the matching head and the regularization term in $\mathcal{L}_{ctr}$}
\label{ablation}
\end{table}

\subsection{Ablation Study}
To evaluate the performance gain of the proposed matching head and the regularization term in $\mathcal{L}_{ctr}$, we conduct ablation experiments on them. According to the experimental results in Table~\ref{ablation}, the proposed matching head results in 3.14\%, 0.90\%, and 5.54\% performance gains on the web, document, and handwriting datasets, respectively. When introducing the regularization term, the proposed method further achieves an improvement of around 1.12\% in average on the four datasets.
% \subsection{Choices of Hyper-parameters}
% \label{choose}
% \noindent\textbf{Choice of $\lambda$} We employ two contrastive losses ($\mathcal{L}_{T}$ and $\mathcal{L}_{I}$) at the training stage of CCR, where $\lambda$ plays an important role in balancing the two loss functions. We conduct experiments with $\lambda$ ranging in \{0, 0.5, 1, 2, 5\}. Through the experimental results (shown in the Supplementary Material), we observe that when the hyperparameter $\lambda$ is set to 1, our method achieves the best performance. Therefore, we set $\lambda$ to 1 for all CCR experiments.

% \noindent\textbf{Choice of $\beta$} To select a appropriate $\beta$, we conduct experiments on the handwriting datasets. When $\beta$ is set to 1, 0.1, 0.01, 0.001, and 0, the performance of our method is 58.76\%, 59.06\%, 59.53\%, 60.30\%, and 59.54\%, respectively. Thus, $\beta$ is set to 0.001 in all CTR experiments.

\section{Discussions}
\label{discussions}

\textbf{Decomposition Levels.} As introduced in Section~\ref{decomposition}, a Chinese character has three types of representations. In the proposed CCR-CLIP model, each type of representation can be fed into the text decoder to extract specific information for each Chinese character. To select the most effective representation for the text encoder, we conduct corresponding experiments. The results reported in Table~\ref{three-level} indicate that the CCR-CLIP model achieves the best performance when the radical-level representation is adopted. The relative poor performance of the stroke-level representation could be attributed to the fact that strokes are too fine-grained to perceive. Therefore, we choose the radical-level representation as the input of the text encoder.

% As introduced in Section~\ref{decomposition}, each Chinese character has three types of representations, \textit{i.e.}, a whole character, a radical sequence, and a stroke sequence. In our method, each kind of representation for Chinese characters can be fed into the text decoder to extract the specific information for each Chinese character. Thus, we conduct experiments to choose the most appropriate representation for the text encoder. The experimental results (see Table~\ref{three-level}) demonstrate that the proposed model achieves the best performance when the radical-level representation is used in the text encoder. A possible reason for the poor performance of stroke-level representation is that strokes are too fine-grained to perceive. Thus, we choose the radical-level representation as the input of the text encoder.

\begin{table}[t]
\renewcommand{\arraystretch}{1.1}
%\begin{subtable}

\centering
\scalebox{0.8}{
\begin{tabular}{c|ccccc}
\hline 
Dataset & Character & Radical & Stroke \\
\hline
HWDB & 96.83\% & \textbf{97.18\%} & 92.74\% \\
CTW & 82.73\% & \textbf{85.78\%} & 83.25\%\\
\hline
\end{tabular}}
\caption{Comparison between different level representations.}
\label{three-level}
\end{table}

\textbf{Visualization.} In order to validate the effectiveness of $\mathcal{L}_{I}$, we sample 1,200 handwritten examples of 12 characters from ICDAR2013~\cite{yin2013icdar} and visualize the embedded visual features in a 2-D space with $t$-SNE, where each character class is denoted by one color. As shown in Figure~\ref{fig:img-loss}(a), some scribbled character samples are far away from the corresponding cluster center in the feature space, which results in incorrect predictions. When $\mathcal{L}_{I}$ is introduced, most of the scribbled character samples are correctly predicted and closer to their cluster centers (see Figure~\ref{fig:img-loss}(b)), which validates the effectiveness of $\mathcal{L}_{I}$ in the proposed CCR-CLIP. More visualization results and failure cases are shown in the Supplementary Material.

\textbf{Limitations.} 
In the proposed method, we incorporate a pre-processing step that the text images are rotated by 90 degrees anticlockwise if they are in a vertical orientation. Since the proposed method is based on canonical representation matching, the features of the same character in different orientations may cause confusion to the model. This may explain why the performance of our method is subpar on the scene dataset.

% \begin{figure}[t]
%     \centering
%     \includegraphics[width=0.47\textwidth]{results.pdf}
%     \caption{Comparison of recognition results in non-zero-shot settings. The red and green characters denote incorrect and correct predictions, respectively.}
%     \label{fig:results}
% \end{figure}

% \begin{figure}[t]
%     \centering
%     \includegraphics[width=0.47\textwidth]{failure_cases.pdf}
%     \caption{Failure cases in non-zero-shot settings. Occluded and scribbled characters still bring difficulties to our method.}
%     \label{fig:failure-case}
% \end{figure}

% \noindent\textbf{Limitations.} 
% Some failure cases in the non-zero-shot settings are visualized in Figure~\ref{fig:failure-case}. In the handwritten settings, scribbled characters are still the primary factor for wrong predictions (see Figure~\ref{fig:failure-case}(left)). The occluded characters in the scene character dataset CTW~\cite{yuan2019large} will also cause difficulties for our method (see Figure~\ref{fig:failure-case}(right)). Since our method relies on the matching between radical sequences and the input character image, and selects the most similar character as the final prediction, occlusions in character images may affect the calculation of similarity.

\section{Conclusion}
In this paper, we propose a novel two-stage framework for Chinese text recognition, which is inspired by the way humans recognize Chinese texts. The first stage involves a CCR-CLIP model that learns canonical representations of Chinese characters by aligning printed character images and Ideographic Description Sequences (IDS). In the second stage, using the learned canonical representations as supervision, we train a Chinese text recognition model with an image-IDS matching head. Extensive experiments demonstrate that the proposed method outperforms previous SOTA methods in both Chinese character recognition and Chinese text recognition tasks.

\section*{Acknowledgements}
This work was supported in part by the National Natural Science Foundation of China (No.62176060), STCSM projects (No.20511100400, No.22511105000), Shanghai Municipal Science and Technology Major Project (No.2021SHZDZX0103), and the Program for Professor of Special Appointment (Eastern Scholar) at Shanghai Institutions of Higher Learning.

{\small
\bibliographystyle{ieee_fullname}
\bibliography{egbib}
}

\appendix
\onecolumn
\section{Choices of Hyperparameters}

In this section, we present the experimental results of determining the appropriate hyperparameters for the proposed CCR-CLIP model. We conduct experiments on the printed artistic character dataset \cite{chen2021zero} for character zero-shot settings and the scene character dataset CTW \cite{yuan2019large} for non-zero-shot settings to choose $\lambda$, and on the handwriting dataset of the CTR benchmark \cite{chen2021benchmarking} to determine $\beta$.

\noindent \textbf{Choice of $\lambda$.}
We use two contrastive losses ($\mathcal{L}_{T}$ and $\mathcal{L}_{I}$) in the training stage of the proposed CCR-CLIP model, and $\lambda$ is the hyperparameter that balances these two loss functions. Table \ref{lambda} shows the experimental results for different values of $\lambda$ ranging from 0 to 5. Based on our experimental results, we find that setting $\lambda$ to 1 achieves the best performance. Furthermore, when $\lambda$ is set to 0, which is the ablation study on $\lambda$, the performance of the CCR-CLIP model is clearly improved with $\lambda=1$, validating the effectiveness of $\mathcal{L}_{I}$. Therefore, we set $\lambda$ to 1 in pre-training experiments.

\noindent \textbf{Choice of $\beta$.} To prevent overfitting on seen characters, we introduce a regularization item in $\mathcal{L}_{tr}$. We conduct experiments on different values of $\beta$ ranging from 0 to 1 and find that the proposed method achieves the highest performance when $\beta$ is set to 0.001 on the CTR benchmark. Specifically, when $\beta$ is set to 0, 0.001, 0.01, 0.1, and 1, the proposed method achieves 59.54\%, 60.30\%, 59.53\%, 59.07\%, and 58.76\%, respectively. Therefore, we set $\beta$ to 0.001 in all experiments on the CTR benchmark.

\begin{table}[ht]
\renewcommand{\arraystretch}{1.1}
%\begin{subtable}
\centering
\scalebox{1.0}{
\begin{tabular}{c|ccccc|c}
\hline 
\multirow{2}*{$\lambda$} &  \multicolumn{5}{c|}{$m$ for Character Zero-Shot Setting}  & \multirow{2}*{CTW}\\
\cline{2-6}
~ & 500 & 1000 & 1500 & 2000 & 2755 &  \\ 
\hline
0 & 23.84\% & 48.13\% & 65.13\% & 72.33\% & 80.48\% & 83.29\% \\
0.5 & 24.49\% & 48.20\% & 65.23\% & 73.55\% & \textbf{81.90\%} & 84.86\%\\
1 & \textbf{25.00\%} & \textbf{49.89\%} & \textbf{65.25\%} & \textbf{74.26\%} & 81.51\% & \textbf{85.78\%} \\
2 & 21.90\% & 48.62\% & 64.96\% & 72.60\% & 81.18\% & 83.12\% \\
5 & 21.42\% & 46.85\% & 61.71\% & 71.60\% & 79.22\% & 83.06\% \\
\hline
\end{tabular}}
\caption{Choice of $\lambda$.}

\label{lambda}
\end{table}

\section{Details of CTR Benchmark}
The CTR benchmark comprises four distinct types of scenarios, namely, scene, web, document, and handwriting. Since the samples of these datasets are collected from various publicly available competitions, projects, and papers, some of the samples may contain non-Chinese characters. Therefore, in this paper, we filtered out such samples as our focus is on Chinese text recognition. Table \ref{details_of_datasets} provides the statistical results of the four filtered datasets. It is worth noting that each of the four datasets includes some zero-shot characters, which pose a significant challenge for existing methods.

\begin{table}[ht]
\renewcommand{\arraystretch}{1.0}
%\begin{subtable}
\centering
\scalebox{1.0}{
\begin{tabular}{cccccc}
\hline 
Dataset & Training & Validation & Test & Alphabet Size & ZS Characters \\
\hline
Scene & 369085 & 45876 & 46062 & 5326 & 103\\
Web &  52103 & 6585 & 6454 & 3843 & 81\\
Document & 158317 & 20025 & 19905 & 4301 & 51\\
Handwriting & 34830 & 8876 & 11018 & 5051 & 227\\
\hline
\end{tabular}}
\caption{The statistical results of four datasets. ``ZS Characters'' represents the number of zero-shot characters in the test dataset.}

\label{details_of_datasets}
\end{table}

\section{Examples of Adopted Datasets}
\begin{figure*}[ht]
    \centering\includegraphics[width=0.6\textwidth]{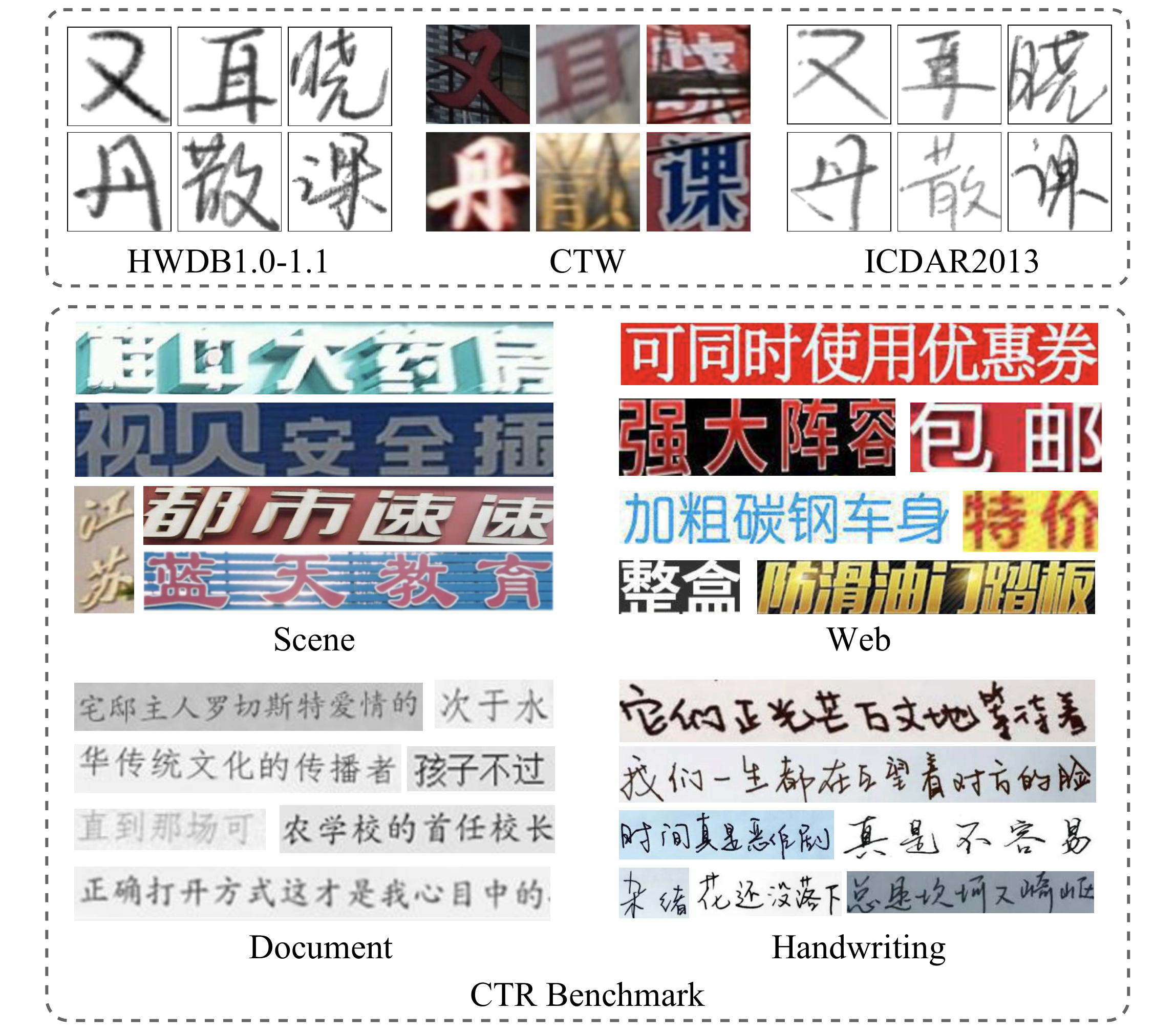}
    \caption{Examples of the adopted datasets.}
    \label{fig:datasets}
\end{figure*}

In this paper, we evaluate the proposed method in Chinese character recognition and Chinese text recognition tasks, where four datasets (\textit{i.e.}, HWDB1.0-1.1 \cite{liu2013online}, ICDAR2013 \cite{yin2013icdar}, CTW \cite{yuan2019large}, and CTR benchmark \cite{chen2021benchmarking}) are adopted. Some examples of these datasets are shown in Figure \ref{fig:datasets}.

\section{More Experimental Results}
In the Chinese character recognition task, we conduct additional zero-shot experiments to evaluate the effectiveness of the proposed CCR-CLIP model. We follow \cite{chen2021zero} to construct corresponding datasets for character zero-shot and radical zero-shot settings. For character zero-shot settings, we collect samples with labels falling in the first $m$ classes as the training set and the last $k$ classes as the test set. For the handwritten character dataset HWDB, $m$ ranges in \{500, 1000, 1500, 2000, 2755\} and $k$ is set to 1000; for the scene character dataset CTW, $m$ ranges in \{500, 1000, 1500, 2000, 3150\} and $k$ is set to 500. For radical zero-shot settings, we first calculate the frequency of each radical in the lexicon. Then the samples of characters that have one or more radicals appearing less than $n$ times are collected as the test set, otherwise, collected as the training set, where $n$ ranges in \{10, 20, 30, 40, 50\} in radical zero-shot settings. It is important to note that even though radicals in the test set may be few-shot, we still use the term ``radical zero-shot setting" in accordance with previous work \cite{chen2021zero}.

The experimental results presented in Table \ref{big-table-supp} demonstrate that the proposed CCR-CLIP model outperforms the compared methods by a clear margin in both character zero-shot and radical zero-shot settings. This improvement can be attributed to the architecture of aligning IDSs and character images, which enables the model to better capture the discriminative features of characters. Furthermore, the introduction of contrastive loss $\mathcal{L}_{I}$ between the input images of the same character helps the feature extractor to focus on the texture of characters rather than complex backgrounds, resulting in further performance improvement. Compared with those methods that introduce template character images during training, the proposed CCR-CLIP model can still achieve the best performance (shown in Table \ref{additional-data}).

\begin{table*}[ht]
\renewcommand{\arraystretch}{1.0}
%\begin{subtable}

\centering
\scalebox{0.80}{
\begin{tabular}{c|ccccc | ccccc }
\hline 
\multirow{2}*{\textbf{HWDB}} &  \multicolumn{5}{c|}{$m$ for Character Zero-Shot Setting} &  \multicolumn{5}{c}{$n$ for Radical Zero-Shot Setting}\\
\cline{2-11}
~ & 500 & 1000 & 1500 & 2000 & 2755 & 50 & 40 & 30 & 20 & 10  \\ 
\hline
DenseRAN \cite{wang2018denseran} & 1.70\% & 8.44\% & 14.71\% & 19.51\% & 30.68\% & 0.21\% & 0.29\% & 0.25\% & 0.42\% & 0.69\% \\
HDE \cite{cao2020zero} & 4.90\% & 12.77\% & 19.25\% & 25.13\% & 33.49\% & 3.26\% & 4.29\% & 6.33\% & 7.64\% & 9.33\%\\
Chen et al. \cite{chen2021zero} & 5.60\% & 13.85\% & 22.88\% & 25.73\% & 37.91\% & 5.28\% & 6.87\% & 9.02\% & 14.67\% & 15.83\%\\
Ours & \textbf{21.79\%} & \textbf{42.99\%} & \textbf{55.86\%} & \textbf{62.99\%} & \textbf{72.98\%} & \textbf{11.15\%} & \textbf{13.85\%} & \textbf{16.01\%} & \textbf{16.76\%} & \textbf{15.96\%}\\
\hline
\hline
\multirow{2}*{\textbf{CTW}} &  \multicolumn{5}{c|}{$m$ for Character Zero-Shot Setting} &  \multicolumn{5}{c}{$n$ for Radical Zero-Shot Setting}\\
\cline{2-11}
~ & 500 & 1000 & 1500 & 2000 & 3150 & 50 & 40 & 30 & 20 & 10  \\ 
\hline
DenseRAN \cite{wang2018denseran} & 0.15\% & 0.54\% & 1.60\% & 1.95\% & 5.39\% & 0\% & 0\% & 0\% & 0\% & 0.04\% \\
HDE \cite{cao2020zero} & 0.82\% & 2.11\% & 3.11\% & 6.96\% & 7.75\% & 0.18\% & 0.27\% & 0.61\% & 0.63\% & 0.90\%\\
Chen et al. \cite{chen2021zero} & 1.54\% & 2.54\% & 4.32\% & 6.82\% & 8.61\% & 0.66\% & 0.75\% & 0.81\% & 0.94\% & 2.25\% \\
Ours & \textbf{3.55\%} & \textbf{7.70\%} & \textbf{9.48\%} & \textbf{17.15\%} & \textbf{24.91\%} & \textbf{0.95\%} & \textbf{1.77\%} & \textbf{2.36\%} & \textbf{2.59\%} & \textbf{4.21\%}\\
\hline
\end{tabular}}
\caption{The experimental results in the character zero-shot settings (left) and radical zero-shot settings (right). $m$ represents that samples of the first $m$ classes are used for training in the character zero-shot settings; $n$ represents that samples with one or more radicals appearing less than $n$ time are collected for testing in the radical zero-shot settings. These experiments do not involve additional template character images during training.}
\label{big-table-supp}
\end{table*}

\begin{table}[ht]
\renewcommand{\arraystretch}{1.0}
%\begin{subtable}

\centering
\scalebox{0.79}{
\begin{tabular}{c|ccccc|ccccc}
\hline 
 ~ & \multicolumn{5}{c|}{$m$ for Character Zero-Shot Setting (\textbf{HWDB})} &  \multicolumn{5}{c}{$m$ for Character Zero-Shot Setting (\textbf{CTW})}\\
\cline{2-11}
~ & 500 & 1000 & 1500 & 2000 & 2755 & 500 & 1000 & 1500 & 2000 & 3150\\ 
\hline
DMN \cite{li2020deep} & 66.33\% & 79.09\% & 84.14\% & 86.79\% & 88.98\% & 0.47\% & 1.20\% & 0.93\% & 1.60\% & 3.12\%\\
CMPL \cite{ao2022cross} & 72.49\% & 80.57\% & 84.40\% & 86.47\% & 89.29\% & - & - & - & - & -\\
CCD \cite{liu2022open} & 90.93\% & 94.10\% & 94.58\% & 95.55\% & - & 58.22\% & 68.56\% & 74.45\% & 77.18\% & - \\
Ours & \textbf{93.80\%} & \textbf{94.97\%} & \textbf{95.35\%} & \textbf{95.71\%} & \textbf{95.73\%} & \textbf{62.13\%} & \textbf{70.16\%} & \textbf{75.88\%} & \textbf{78.85\%} & \textbf{80.03\%} \\
\hline
\end{tabular}}
\caption{Comparison with previous methods in the case of using template character images during training.}
\label{additional-data}
\end{table}

\section{Visualizations of Recognition Results and Failure Cases}
\begin{figure*}[ht]
    \centering\includegraphics[width=0.8\textwidth]{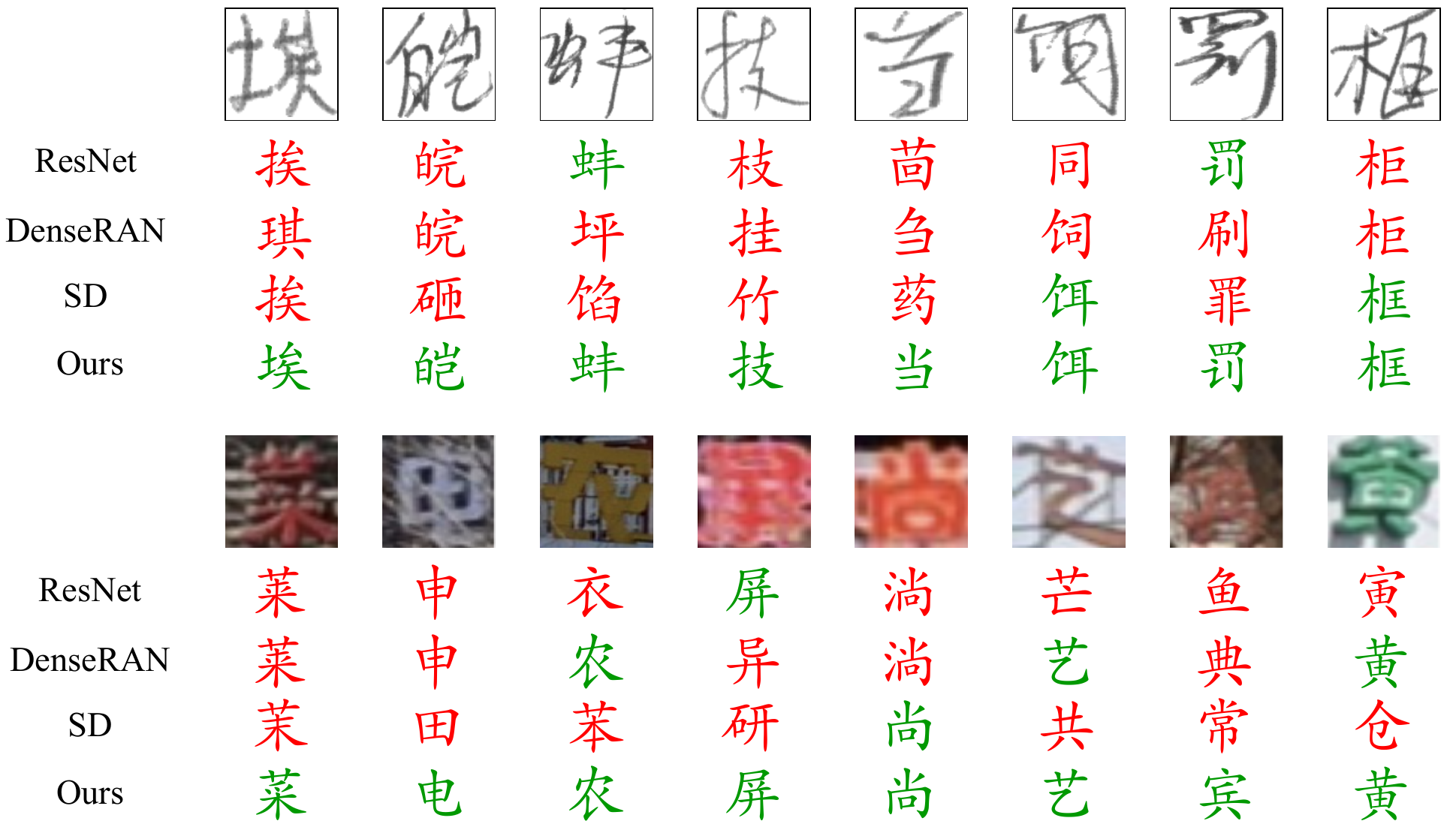}
    \caption{Recognition results of CCR.}
    \label{ccr-res}
\end{figure*}

\begin{figure*}[ht]
    \centering\includegraphics[width=0.8\textwidth]{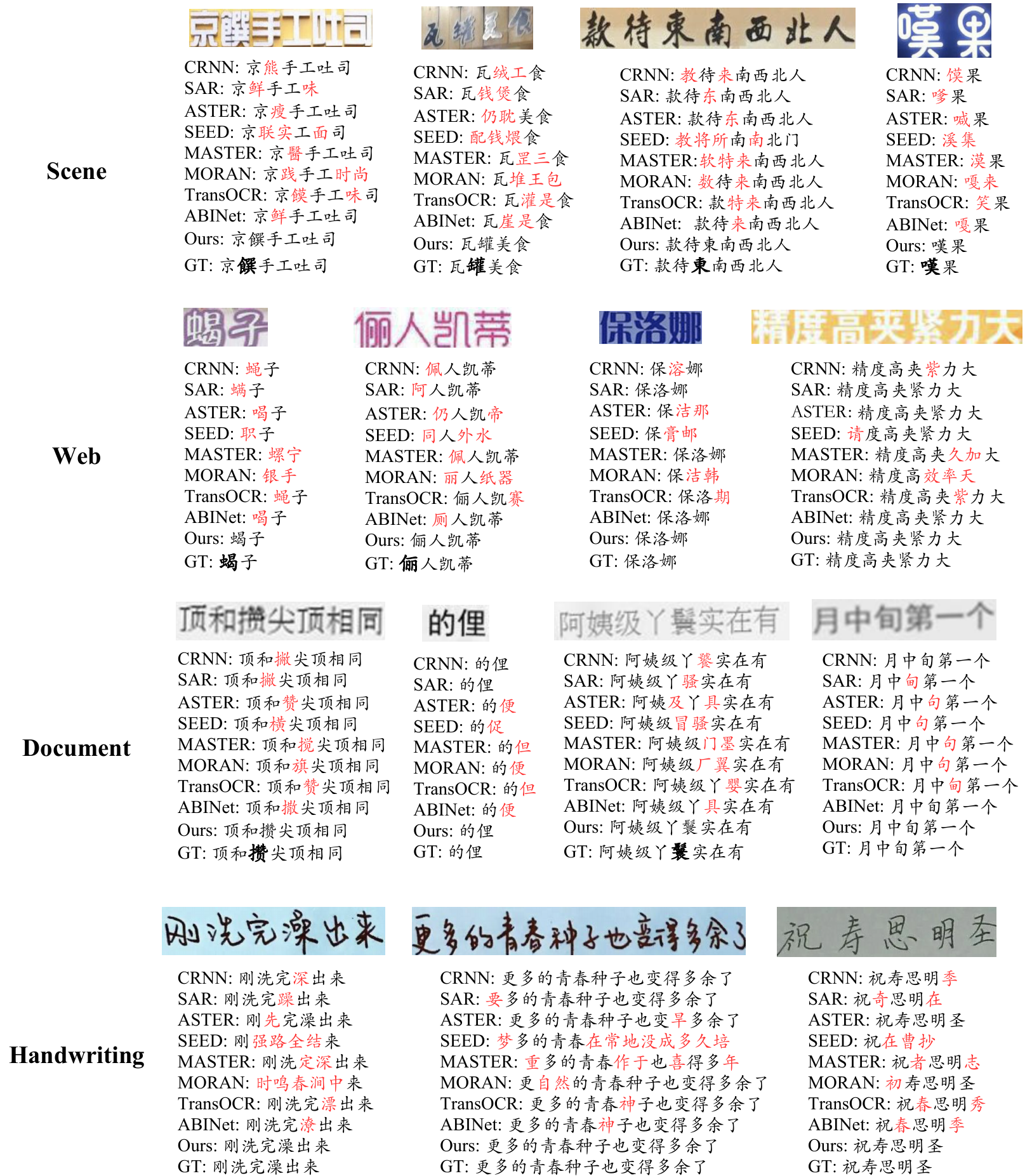}
    \caption{Recognition results of CTR. Red characters indicate wrongly predicted results, while bold characters represent zero-shot and few-shot ones in the training dataset.}
    \label{ctr-res}
\end{figure*}

In this section, we visualize some recognition results of the proposed method including results of CCR and CTR. Compared with decompose-based methods \cite{chen2021zero, wang2018denseran}, the proposed CCR-CLIP model is more robust to the characters with scribbled strokes and complex backgrounds in the non-zero-shot setting, which benefits from the utilization of loss $\mathcal{L}_I$ between character images with the same label (shown in Figure \ref{ccr-res}). Additionally, we evaluate the proposed method on the CTR task and demonstrate its superior performance in recognizing zero-shot and few-shot Chinese characters, as shown in Figure \ref{ctr-res}.

\begin{figure*}[ht]
    \centering\includegraphics[width=0.8\textwidth]{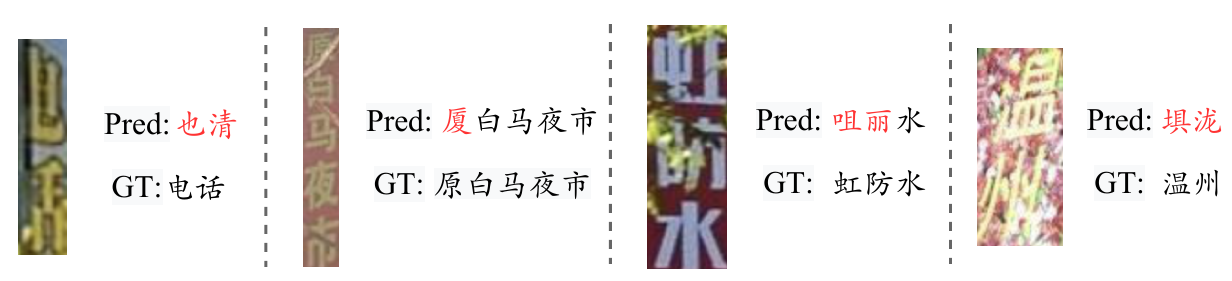}
    \caption{Visualizations of failure cases.}
    \label{failure-case}
\end{figure*}

As mentioned in the main text, the proposed method includes a pre-processing step where text images are rotated by 90 degrees anticlockwise if they are in a vertical orientation. Visualizations of failure cases shown in Figure \ref{failure-case} demonstrate that features of the same character in different orientations may cause confusion in the proposed model because it relies on canonical representation matching.

% \newpage
% \newpage
% {\small
% \bibliographystyle{ieee_fullname}
% \bibliography{egbib}
% }

\end{document}